\documentclass[runningheads]{llncs}
\usepackage[T1]{fontenc}
\usepackage{graphicx}
\usepackage{ upgreek }
\usepackage{multirow}
\usepackage{graphicx}
\usepackage{xspace}
\usepackage{amsmath}
\usepackage{amsfonts}
\usepackage[table,xcdraw]{xcolor}

\def\PCN{PCN\@\cite{yuan2018pcn}\xspace}
\def\ECG{ECG\@\cite{pan2020ecg}\xspace}
\def\VRCnet{VRCnet\@\cite{pan2021vrcn}\xspace}
\def\SFN{SFN\@\cite{xiang2021snowflakenet}\xspace}
\def\PointAttN{PointAttN\@\cite{wang2022pointattn}\xspace}

\begin{document}
%
% \title{Contribution Title\thanks{Supported by organization x.}}
% \title{Point2SSM: Unsupervised Learning of Anatomic Statistical Shape Models from Point Clouds}
%\title{Can point cloud deep learning provide unsupervised anatomical statistical shape models?}
\title{Can point cloud networks learn statistical shape models of anatomies?}
%

% \author{Anonymous}

\author{Jadie Adams\inst{1,2} \and
Shireen Elhabian\inst{1,2}}
%index{Adams, Jadie}
%index{Elhabian, Shireen}
%
\authorrunning{Adams and Elhabian}
\institute{Scientific Computing and Imaging Institute, University of Utah, UT, USA \and
School of Computing, University of Utah, UT, USA \\
\email{ jadie.adams@utah.edu, shireen@sci.utah.edu }
}

\maketitle  

\begin{abstract}
\setcounter{footnote}{0}
Statistical Shape Modeling (SSM) is a valuable tool for investigating and quantifying anatomical variations within populations of anatomies. 
% However, traditional correspondence-based SSM generation methods require a time-consuming re-optimization process each time a new subject is added to the cohort, making the inference process prohibitive for clinical research. 
However, traditional correspondence-based SSM generation methods have a prohibitive inference process and require complete geometric proxies (e.g., high-resolution binary volumes or surface meshes) as input shapes to construct the SSM. Unordered 3D point cloud representations of shapes are more easily acquired from various medical imaging practices (e.g., thresholded images and surface scanning). Point cloud deep networks have recently achieved remarkable success in learning permutation-invariant features for different point cloud tasks (e.g., completion, semantic segmentation, classification). However, their application to learning SSM from point clouds is to-date unexplored. In this work, we demonstrate that existing point cloud encoder-decoder-based completion networks can provide an untapped potential for SSM, capturing population-level statistical representations of shapes while reducing the inference burden and relaxing the input requirement. We discuss the limitations of these techniques to the SSM application and suggest future improvements. Our work paves the way for further exploration of point cloud deep learning for SSM, a promising avenue for advancing shape analysis literature and broadening SSM to diverse use cases.

\end{abstract}
\section{Introduction}
Statistical Shape Modeling (SSM) enables population-based morphological analysis, which can reveal patterns and correlations between shape variations and clinical outcomes. 
SSM can help researchers understand the differences between healthy and pathological anatomy, assess the effectiveness of treatments, and identify biomarkers for diseases  
(e.g., \cite{atkins2017quantitative,bischoff2014incorporating,merle2019high,carriere2014apathy,bruse2016statistical}).
The traditional pipeline for constructing SSM entails the segmentation of 3D images 
to acquire either binary volumes or meshes and then aligning these shapes.
SSM can then be constructed \textit{explicitly} via finding surface-to-surface correspondences across the cohort or \textit{implicitly} via deforming a predefined 
atlas to each shape sample. 
Correspondence-based shape models are widely used due to their intuitive interpretation \cite{sarkalkan2014statistical}; they comprise of sets of landmarks or \textit{correspondence points} that are defined consistently using invariant points across populations that vary in their form.
Historically, correspondence points were established manually to capture biologically significant features.
This cumbersome, subjective process has since been replaced via automated \textit{optimization}, which defines dense sets of correspondence points, aka a Point Distribution Model (PDM). 
% Optimization schemes
PDM optimization schemes have been defined using metrics such as entropy and minimum description length \cite{cates2007shape,davies2002minimum} and via parametric representations \cite{ovsjanikov2012functional,styner2006framework}.
% Inference limitations
A significant drawback of these methods is that PDM optimization must be performed on the entire shape cohort of interest, which is time-consuming and hinders inference. 
To evaluate a new patient scan using PDM, the new scan must undergo segmentation and alignment, and the PDMs must be optimized again for the entire population of shapes.
%the shape must be segmented, aligned, and optimized again for the entire population of shapes. 
% This overhead has prevented the widespread adoption of SSM in medical research.
% input limitation
Moreover, current approaches require a complete, high-resolution mesh or binary volume representation of the shape that is free from noise and artifacts. Therefore, lightweight shape acquisition methods (such as thresholding clinical images, anatomical surface scanning, and shape acquired from stacked or orthogonal 2D slices) cannot be directly used for SSM \cite{timmins2021effect,3Dscanning}.
% Deep learning
Deep learning solutions have been proposed to mitigate these limitations by predicting PDMs directly from unsegmented 3D images using convolutional neural networks \cite{bhalodia2018deepssm,uncertaindeepssm,adams2022images}.
% Limitations
However, these frameworks require PDM supervision and, hence, need the traditional optimization-based workflow to acquire training data. 

Effective SSM from point clouds would be widely applicable in clinical research, from artery disease progression from point clouds acquired via biplane angiographic and intravascular ultrasound image data fusion \cite{timmins2021effect}, to orthopedics implant design from point clouds acquired via 3D body scanning \cite{3Dscanning}. Applying existing methods for SSM generation from point clouds requires converting them to meshes or rasterizing them into segmentations, which is nontrivial given that point clouds are unordered and do not retain surface normals.  
SSM directly from point clouds would enable many clinical studies. 
Recently point cloud deep learning has gained attention, with significant efforts focused on effective point completion networks (PCNs) for generating complete point clouds from partial observations.
% , which is crucial for many downstream tasks.
Most point-completion methods use order-invariant feature encoding and two-stage coarse-to-fine decoding.
% PCNs do SSM
An important outcome of such networks, which has been overlooked and not reported, is that the learned coarse point clouds are ordered and provide correspondence.
% contribution
In this work, we acknowledge this missed potential and explore the use of PCNs to predict PDM from 3D point clouds in an unsupervised manner.
We investigate state-of-the-art PCNs as potential solutions for generating PDMs that (1) accurately represent shapes via uniformly distributed points constrained to the surface and (2) provide good correspondences that capture population-level statistics.\footnote{Source code is available at: \url{https://github.com/jadie1/PointCompletionSSM}}
We discuss the benefits of this approach, its robustness to missingness and training size, current limitations, and possible improvements.
This discussion will bring awareness to the community about the potential for learning SSM from point clouds and ultimately make SSM a more accessible, viable option in future clinical research.

\section{Background}
\subsubsection{Point Distribution Models}
The goal of SSM is to capture the inherent characteristics or underlying parameters of a shape class that remain when global geometrical information is removed. 
Given a PDM,  correspondence points can be averaged across subjects to provide a mean shape, and principal component analysis (PCA) can be performed to compute the modes of shape variation, which can then be visualized and used in downstream medical tasks. 
Furthermore, if a PDM contains sub-populations, such as disease versus control, the differences in mean shapes can be quantified and visualized, providing group characterization. 

\subsubsection{Point Cloud Deep Learning}
Deep learning from 3D point clouds is an emerging area of research with numerous applications in computer vision, robotics, and medicine (e.g., classification, object tracking, segmentation, registration, pose estimation) \cite{akagic2022pointcloudsurvey}. 
PointNet \cite{qi2017pointnet} pioneered a Multi-Layer Perceptron (MLP) and max-pooling-based approach for permutation invariant feature learning from raw point clouds. 
FoldingNet \cite{yang2018foldingnet} proposed a point cloud auto-encoder with a folding-based decoder that utilizes 2D grid deformation for reconstruction. 
Several convolutional approaches have been proposed, including mapping point clouds to voxel grids to directly apply 3D CNNs \cite{le2018pointgrid} and graph-based methods \cite{wang2019dgcnn}.

\begin{figure}
\begin{centering}
    \includegraphics[width=.8\textwidth]{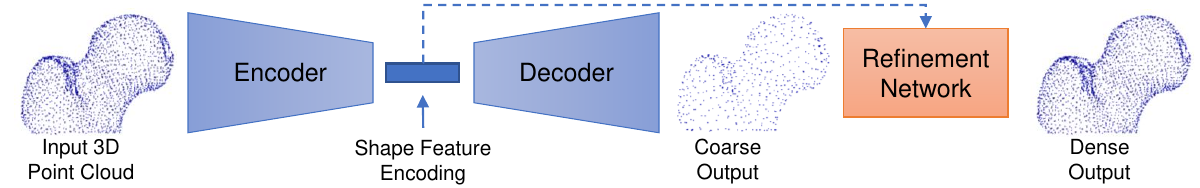}
    \caption{General coarse-to-fine architecture of the considered PCNs.}
    \label{fig:arch}
\end{centering}
\end{figure}

The initial \textit{point cloud completion} network, PCN \cite{yuan2018pcn}, utilized PointNet \cite{qi2017pointnet} and FoldingNet \cite{yang2018foldingnet} with a coarse-to-fine decoder.
Since then, numerous point cloud completion approaches have been proposed, including point MLP and folding-based extensions, 3D convolution approaches, graph-based methods, generative modeling approaches (including generative adversarial network GAN-based and variational autoencoder VAE-based), and transformer-based methods. 
See \cite{fei2022pcnsurvey} for a recent survey.
Many approaches follow the general framework of first encoding the point cloud into a permutation-invariant feature representation, then decoding the encoded shape feature to acquire a coarse or sparse point cloud, 
and finally, refining the coarse point cloud to acquire the dense complete prediction. The general architecture of these methods is shown in Figure \ref{fig:arch}.

\section{Methods}
\subsection{Point Completion Networks for SSM}
Our experiments demonstrate that when coarse-to-fine point completion networks are trained on anatomical shapes, the bottleneck captures a population-specific shape prior. 
Directly decoding the shape feature representation results in a consistent ordering of the intermediate coarse point cloud across samples, providing a PDM. This phenomenon can be intuitively understood as an application of Occam's razor, where the model prefers to learn the simplest solution, resulting in consistent output ordering. Many point completion networks contain skip connections from the feature and/or input space to the refinement network. In the case where the unordered input point cloud is fed to the refinement network, the ordering of the output dense point cloud is understandably lost. 

To study whether point completion networks can learn anatomical SSM, we first extract point clouds from mesh vertices, %preprocess input point clouds from meshes, 
then train point completion models, and finally evaluate the effectiveness of the predicted coarse point clouds as PDMs. Note this approach is not restricted to input point clouds obtained from meshes; point clouds from any acquisition process can be used. 
Global geometric information is factored out by aligning all shapes via iterative closest points \cite{besl1992method} to a reference shape.
We utilized the open-source toolkit ShapeWorks \cite{cates2017shapeworks} for this step. In our experiments, the aligned, unordered mesh vertices serve as ground truth complete point clouds. The ground truth points are randomly downsampled to 2048 points and permuted to provide input point clouds. As is standard, the point clouds are uniformly scaled to be between -1 and 1 to assist network training. 
We consider a state-of-the-art model from the major point completion approach categories: 
\begin{itemize}
    \item \textbf{\PCN}: Point Completion Network (MLP-based)
    \item\textbf{\ECG}: Edge-aware Point Cloud Completion with Graph Convolution (convolution-based)
    \item\textbf{\VRCnet}: Variational Relational Point Completion Network (generative-modeling based)
    \item\textbf{\SFN}: SnowflakeNet (transformer-based)
    \item\textbf{\PointAttN}: Point Attention Network (attention-based)   
\end{itemize}   

Point completion network loss is based on the \textit{L1} \textbf{Chamfer Distance (CD)} \cite{fei2022pcnsurvey}, which defines the minimum distance between two sets of points. 
The loss is typically defined as a combination of coarse and dense loss with weighting parameter $\alpha$, which we consistently set across models. All model's hyperparameters are set to the original implementation values, and training is run until convergence (as assessed by training CD).

%%%%%%%%%%%%%%%%%%%%%%%%%%%%%%%%%%%%%%%%%%%%%%%%%%%%%%%%%%
\subsection{Evaluation Metrics}
In addition to CD, \textbf{Fscore} \cite{tatarchenko2019fscore} is typically used in point completion to quantify the percentage of points that are reconstructed correctly. 
In analyzing accurate surface sampling for SSM, we quantify the \textbf{point-to-face distance (P2F)} from each point in the predicted point cloud to the closest surface of the ground truth mesh. 
To analyze point \textbf{uniformity}, we quantify the variance in the distance of each point to its six nearest neighbors. A uniform PDM would result in a small variance in the point nearest neighbor distance.

We also consider PDM correspondence analysis metrics. An ideal PDM is compact, meaning that it represents the distribution of the training data using the minimum number of parameters. We quantify \textbf{compactness} as the number of PCA modes required to capture 99\% of the variation in the correspondences.
% A PDM with fewer modes is considered more compact. 
A good PDM should also generalize well from training examples to unseen examples.
The \textbf{generalization} metric is defined as the reconstruction error ($L2$) between  predicted correspondences of a held-out point cloud and the correspondences reconstructed via the training PDM.
Finally, effective SSM is specific, generating only valid instances of the shape class in the training set. 
The average distance between correspondences sampled from the training PDM and the closest existing training correspondences provides the \textbf{specificity} metric.
\section{Experiments}

We use five datasets in experiments -- one synthetic ellipsoid dataset for a proof-of-concept and four real anatomical shapes: proximal femurs, left atrium of the heart, spleen, and pancreas. These datasets vary greatly in size (see Table \ref{table:results}, Column 1) and shape variation. Details and visualization of these cohorts are provided in the supplementary materials. 
In all experiments, the input point cloud size is 2048, the coarse output size is 512, and the dense output size is 2048.
The real datasets were split (90\%/10\%) into training and testing sets. Stratified sampling via clustering was used to define the test set to ensure it is representative, given the low sample size. 

%%%%%%%%%%%%%%%%%%%%%%%%%%%%%%%%%%%%%%%%%%%%%%%%%%%%%%%%%%%%%%%%%%%%%%%%%%%%%%%%
\begin{table}[!h]
\caption{\textbf{Results:} Evaluation metrics across all models and datasets. All metrics are quantified on coarse predictions, with the exception of the gray column, which reports the dense prediction CD. The CD values are scaled by 1000 for reporting and Fscore is calculated at 1\% threshold. Best test values are shown in bold.}
\label{table:results}
\resizebox{\textwidth}{!}{%
\begin{tabular}{|c|c|ccccc|ccc|}
\hline
 &  & \multicolumn{5}{c|}{\textbf{Point Accuracy Metrics (train/test)}} & \multicolumn{3}{c|}{\textbf{SSM Evaluation Metrics}} \\ \cline{3-10} 
\multirow{-2}{*}{\textbf{Data}} & \multirow{-2}{*}{\textbf{Model}} & \cellcolor[HTML]{C0C0C0}\textbf{Dense CD $\downarrow$} & \textbf{CD $\downarrow$} & \textbf{Fscore $\uparrow$} & \textbf{P2F (mm) $\downarrow$} & \textbf{Uniformity $\downarrow$} & \textbf{Comp.$\downarrow$} & \textbf{Gen. $\downarrow$} & \textbf{Spec.$\downarrow$} \\ \hline
 % ellipsoids
& \PCN & \cellcolor[HTML]{E7E6E6}0.908/0.917 & 1.80/1.80 & 0.561/\textbf{0.582} & 0.0142/\textbf{0.0168} & \multicolumn{1}{c|}{0.325/0.354} & \textbf{2} & 0.140 & \multicolumn{1}{c|}{0.348} \\
 & \ECG & \cellcolor[HTML]{E7E6E6}0.929/0.923 & 1.80/1.78 & 0.554/0.579 & 0.0171/0.0196 & \multicolumn{1}{c|}{0.308/0.335} & \textbf{2} & 0.136 & \multicolumn{1}{c|}{0.348} \\
 & \VRCnet & \cellcolor[HTML]{E7E6E6}0.923/0.919 & 1.78/\textbf{1.77} & 0.563/\textbf{0.582} & 0.0384/0.0367 & \multicolumn{1}{c|}{0.295/0.327} & \textbf{2} & 0.143 & \multicolumn{1}{c|}{0.348} \\
 & \SFN & \cellcolor[HTML]{E7E6E6}0.842/\textbf{0.838} & 1.92/1.91 & 0.559/0.566 & 0.0960/0.0949 & \multicolumn{1}{c|}{0.237/0.256} & \textbf{2} &\textbf{ 0.103 }& \multicolumn{1}{c|}{\textbf{0.332}} \\
\multirow{-5}{*}{\begin{tabular}[c]{@{}c@{}}\textbf{Ellipsoids}\\ \\ Train: 50\\ Test: 30\end{tabular}} & \PointAttN & \cellcolor[HTML]{E7E6E6}0.912/0.911 & 1.78/\textbf{1.77} & 0.554/0.571 & 0.0255/0.0315 & \multicolumn{1}{c|}{0.217/\textbf{0.243}} & \textbf{2} & 0.137 & \multicolumn{1}{c|}{0.356} \\ \hline
% femur
 & \PCN & \cellcolor[HTML]{E7E6E6}1.27/1.80 & 2.30/\textbf{2.91} & 0.501/\textbf{0.386} & 0.235/\textbf{0.712} & 1.77/1.63 & 18 & 0.28 & 1.30 \\
 & \ECG & \cellcolor[HTML]{E7E6E6}1.59/2.29 & 2.37/3.02 & 0.485/0.377 & 0.297/0.750 & 1.71/1.63 & 15 & 0.255 & 1.23 \\
 & \VRCnet & \cellcolor[HTML]{E7E6E6}1.78/2.33 & 2.88/3.27 & 0.399/0.362 & 0.676/0.876 & 1.62/1.60 & \textbf{5} & 0.259 & 0.786 \\
 & \SFN & \cellcolor[HTML]{E7E6E6}1.04/\textbf{1.29} & 2.36/3.13 & 0.464/0.365 & 0.341/0.774 & 1.01/\textbf{0.967} & 17 & 0.297 & 1.37 \\
\multirow{-5}{*}{\begin{tabular}[c]{@{}c@{}}\textbf{Femur}\\ \\ Train: 51\\ Test: 5\end{tabular}} & \PointAttN & \cellcolor[HTML]{E7E6E6}1.27/1.50 & 3.25/3.94 & 0.352/0.295 & 0.759/1.03 & 2.32/2.29 & 9 & \textbf{0.199} & \textbf{0.905} \\ \hline
 % left atrium
 & \PCN & \cellcolor[HTML]{E7E6E6}0.405/0.773 & 0.707/1.09 & 0.941/0.865 & 0.245/1.02 & 1.87/2.01 & 82 & 0.932 & 5.85 \\
 & \ECG & \cellcolor[HTML]{E7E6E6}0.571/0.868 & 0.814/1.16 & 0.919/0.848 & 0.440/1.07 & 1.85/2.04 & 57 & 0.929 & 5.22 \\
 & \VRCnet & \cellcolor[HTML]{E7E6E6}0.070/\textbf{0.085} & 0.886/1.62 & 0.904/\textbf{0.768} & 0.632/1.53 & 1.87/2.19 & 81 & 1.04 & 6.09 \\
 & \SFN & \cellcolor[HTML]{E7E6E6}0.310/0.311 & 0.822/\textbf{0.948} & 0.927/0.902 & 0.511/\textbf{0.831} & 1.30/\textbf{1.46} & \textbf{49} & \textbf{0.783} & \textbf{5.07} \\
\multirow{-5}{*}{\begin{tabular}[c]{@{}c@{}}\textbf{Left Atrium}\\ \\ Train: 987\\ Test: 109\end{tabular}} & \PointAttN & \cellcolor[HTML]{E7E6E6}0.332/0.360 & 0.875/1.20 & 0.909/0.840 & 0.523/1.08 & 1.65/1.79 & 82 & 0.807 & 5.46 \\ \hline
% Spleen
 & \PCN & \cellcolor[HTML]{E7E6E6}3.59/7.88 & 4.51/9.77 & 0.326/0.155 & 1.67/3.73 & 15.3/12.0 & 7 & 1.47 & 4 \\
 & \ECG & \cellcolor[HTML]{E7E6E6}1.84/6.46 & 3.0/9.69 & 0.456/0.174 & 1.07/3.94 & 10.2/7.98 & 14 & 1.62 & 5.57 \\
 & \VRCnet & \cellcolor[HTML]{E7E6E6}0.408/\textbf{0.583} & 5.03/14.3 & 0.318/\textbf{0.117} & 1.81/4.59 & 16.1/12.5 & 6 & 1.73 & 4.48 \\
 & \SFN & \cellcolor[HTML]{E7E6E6}0.986/1.28 & 4.57/\textbf{7.23} & 0.265/0.189 & 1.64/\textbf{3.07} & 17.8/14.7 & \textbf{6} & \textbf{1.08} & \textbf{4.18} \\
\multirow{-5}{*}{\begin{tabular}[c]{@{}c@{}}\textbf{Spleen}\\ \\ Train: 36\\ Test: 4\end{tabular}} & \PointAttN & \cellcolor[HTML]{E7E6E6}1.13/3.70 & 3.49/11.5 & 0.380/0.158 & 1.33/4.22 & 7.67/\textbf{6.72} & 15 & 1.86 & 6.52 \\ \hline

 % Pancreas
 & \PCN & \cellcolor[HTML]{E7E6E6}0.571/1.63 & 0.869/2.02 & 0.895/0.710 & 0.526/1.92 & 3.44/3.99 & 66 & 1.12 & 5.31 \\
 & \ECG & \cellcolor[HTML]{E7E6E6}1.11/2.51 & 1.10/2.08 & 0.843/0.700 & 0.883/1.99 & 3.53/4.02 & 48 & 1.01 & 4.7 \\
 & \VRCnet & \cellcolor[HTML]{E7E6E6}0.100/\textbf{0.138} & 2.40/3.79 & 0.614/\textbf{0.507} & 2.13/3.17 & 4.75/5.25 & \textbf{18} & \textbf{0.904} & \textbf{3.49} \\
 & \SFN & \cellcolor[HTML]{E7E6E6}0.329/0.557 & 0.95/\textbf{1.85} & 0.880/0.736 & 0.764/\textbf{1.83} & 3.34/\textbf{\textbf{3.61}} & 55 & 0.955 & 5.16 \\
\multirow{-5}{*}{\begin{tabular}[c]{@{}c@{}}\textbf{Pancreas}\\ \\ Train: 245\\ Test: 28\end{tabular}} & \PointAttN & \cellcolor[HTML]{E7E6E6}0.407/0.837 & 1.07/2.55 & 0.849/0.629 & 0.897/2.31 & 3.52/3.65 & 96 & 1.05 & 5.55 \\ \hline
\end{tabular}%
}
\end{table}

%%%%%%%%%%%%%%%%%%%%%%%%%%%%%%%%%%%%%%
\subsubsection{Proof-of-Concept: Ellipsoids}
\begin{figure}[h!]
\centering
\includegraphics[width=.9\textwidth]{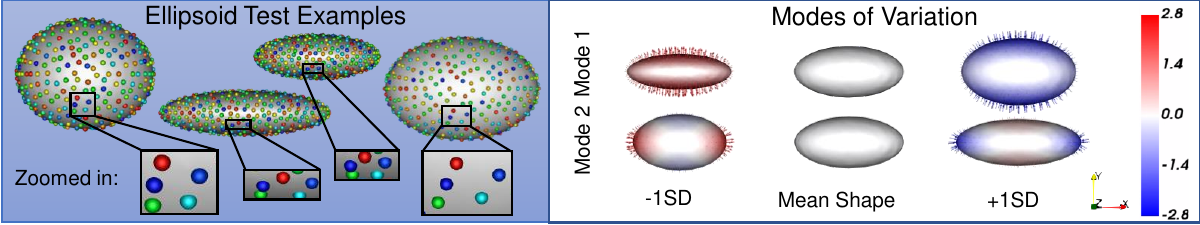}
\caption{\textbf{Ellipsoid PDM from \PCN} Left: Example test predictions on ground truth mesh with color-denoting point correspondence. Points are reasonably uniformly spread with good correspondence, shown via zoomed-in boxes. Right: The primary and secondary modes (training and testing combined) with $\pm1$ standard deviation from the mean shape. Color and vectors denote the difference from the mean. The shape model correctly characterizes the x and y-radius as the only source of variation.} \label{fig:ellipsoids}
\end{figure}

As a proof-of-concept, we generate 3D axis-aligned ellipsoid shapes with fixed $z$-radius and random $x$ and $y$-radius. A testing set of 30 and a training set of just 50 were randomly defined to emulate the scarce data scenario. The results in Table \ref{table:results} show that all of the model variants performed well with low S2F distance (<0.1mm), and all PDMs correctly captured just two modes of variation (x and y-radius). 
Results from the \PCN model are shown in Figure \ref{fig:ellipsoids}.

\subsubsection{Femur}
\begin{figure}[h!]
\centering
\includegraphics[width=.83\textwidth]{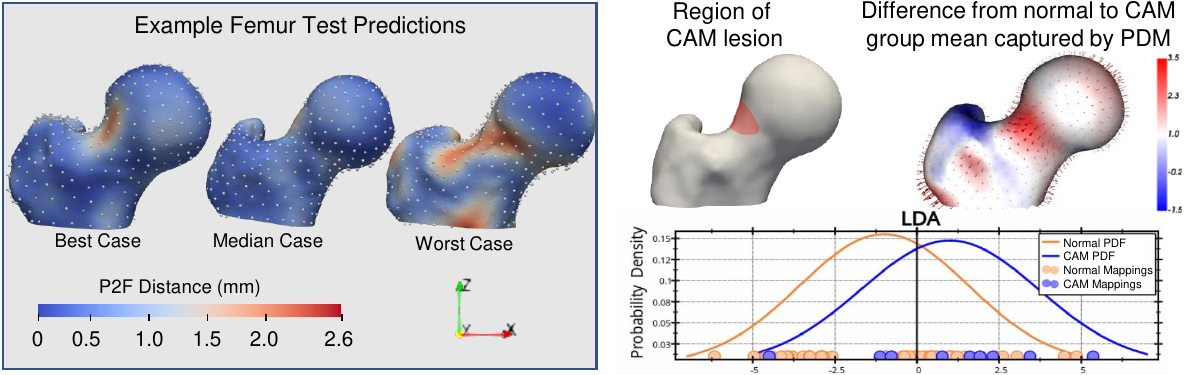}
\caption{\textbf{Femur PDM from \PCN} Left: Test examples with P2F distance displayed as a heatmap. Right: The expected region of CAM lesion correlates with the difference found from the normal to CAM group means.} 
\label{fig:femur}
\end{figure}

The femur dataset is comprised of 56 femoral heads segmented from CT scans, nine of which have the cam-FAI pathology characterized by an abnormal bone growth lesion that causes hip osteoarthritis \cite{atkins2017quantitative}. We utilize this pathology to analyze if PDM from point completion networks can correctly characterize group differences. 
Table \ref{table:results} shows the predicted coarse particles are close to the surfaces (P2F distance of ~0.1mm), and the \PCN model performs best in this regard. 
The \PCN predictions were used to analyze the difference between the normal and CAM pathology mean shapes. Figure \ref{fig:femur} shows the pathology is correctly characterized, and the \textit{Linear Discrimination of Variation} (LDA) plot shows the difference in normal and CAM distributions captured by the PDM. 

\subsubsection{Left Atrium}
\begin{figure}
\centering
\includegraphics[width=.85\textwidth]{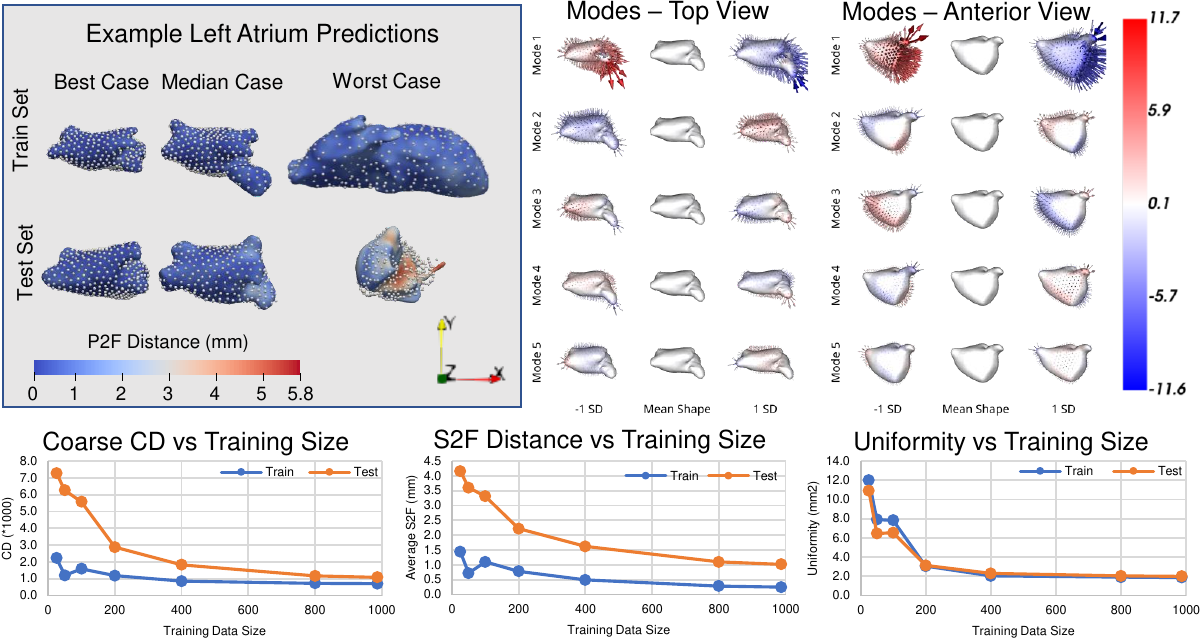}
\caption{Top: Left Atrium PDM from \SFN Left: Train and test examples shown from top view with P2F distance. Right: The first 5 modes of variation from top and anterior view. Bottom: Left Atrium \PCN results vs Training Size}\label{fig:LA}
\vspace{-.1in}
\end{figure}

The left atrium dataset comprises of 1096 shapes segmented from cardiac LGE MRI images from unique atrial fibrillation patients. This cohort contains significant morphological variation in overall size, the size of the left atrium appendage, and the number and arrangement of the pulmonary veins. This variation is reflected in the large compactness values in Table \ref{table:results}. Despite this variation, the models achieve reasonable CD and P2F scores due to the large training size. Figure \ref{fig:LA} shows the PDM predicted by \SFN, which performed best. From the prediction examples, we can see the model represents the training data very well, even in the worst case, which has an extremely enlarged left atrium appendage. The size of the appendage is appropriately captured in the modes of variation and the performance on the test examples is reasonable with the exception of those that are not well represented by the training data, such as the test set worst case. This highlights the importance of a large, representative training set. To further illustrate this importance, we perform an ablation experiment evaluating the performance of the \PCN model with respect to the left atrium training test size (Figure \ref{fig:LA}). 

\subsubsection{Pancreas}
\begin{figure}[h!]
\centering
\includegraphics[width=.9\textwidth]{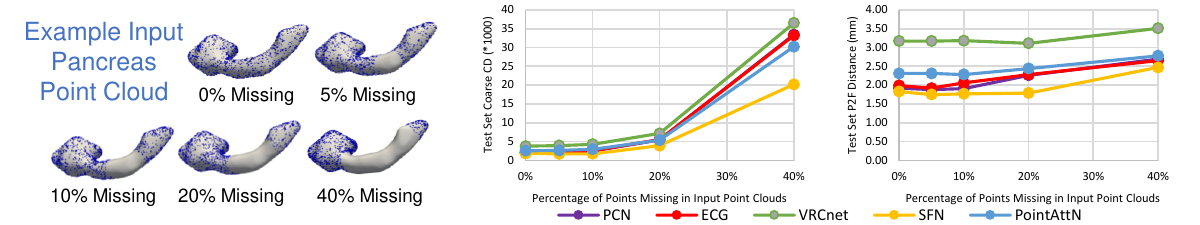}
\caption{\textbf{Effect of Partial Input on Pancreas Prediction Accuracy}}
\label{fig:pancreas}
\end{figure}

We utilize the pancreas dataset \cite{simpson2019medseg} to analyze the impact of incomplete input point clouds as the point completion networks were designed to address. 
Cases of incomplete observations frequently arise in clinical research. For example, in the analysis of bones where some are clipped due to scan size or in cases where 3D shape is interpreted from stacked or orthogonal 2D observations. Traditional methods of SSM generation are unable to handle such cases, but the point cloud learning approach has the potential to. Figure \ref{fig:pancreas} shows how the test set error increase as the percentage of missing points increases.
The \SFN model provides the best results given partial input.

\subsubsection{Spleen}
The spleen dataset \cite{simpson2019medseg} is included to provide an example of a small dataset with a large amount of variation to stress test the point completion models. Table \ref{table:results} shows the models perform the worst on this dataset with regards to CD, Fscore, P2F, and Uniformity. This example illustrates the limitations of this approach to SSM generation. 

\section{Discussion and Conclusion}
Our experiments demonstrate that point cloud networks can learn accurate SSM of anatomy when provided with a sufficiently large and representative training dataset.
The transformer-based \SFN architecture provided the best overall results among the models we explored.
\SFN utilizes k-Nearest Neighbors (kNN) to capture geometric relations in the point cloud, while \PointAttN does not. 
\PointAttN has been shown to provide better point completion of complex man-made objects where kNN information could be misleading. However, in the case of anatomical SSM, it is likely that the kNN information assisted \SFN performance by providing accurate spatial information, given the more convex shape of organs and bones.
Interestingly, the simplest model, \PCN, achieved similar, and sometimes better, SSM accuracy than more current state-of-the-art methods, despite its inferior performance in point completion benchmarks \cite{pan2020ecg,pan2021vrcn,xiang2021snowflakenet,wang2022pointattn}. This may be attributed to point completion benchmarks involving multiple object class point completion, which is a more challenging task. 
Another significant difference between our experiments and point completion benchmarks is the PCN datasets have tens of thousands of examples \cite{yuan2018pcn,pan2021vrcn,yu2021pointr}, while we worked with limited training data -- the typical scenario in shape analysis. 
Our experiments demonstrate that PCNs can effectively predict SSM under limited training data when shape variation is minimal, as in the case of proximal femurs. However, they struggle when there is significant shape variation, such as in the spleen cohort.

This work indicates promising potential for adapting characteristics of point completion architectures and learning schemes to tailor to the task of predicting SSMs from point clouds.  
Potential improvements could be made to the training objective, such as penalization for non-uniformity and bottleneck regularization for compact population-statistical learning. 
Additionally, improvements could be made to address the scarce training data scenario, such as model-based data augmentation and probabilistic transfer learning. 
Although we evaluated only smooth point cloud inputs, similar architectures have shown success in point cloud denoising tasks, suggesting that our approach may handle noise as well \cite{alliegro2021denoise,sahin2022cmd}. 
This work establishes the groundwork for future research into the potential of point cloud deep learning for SSM, offering significant benefits over traditional SSM generation, including: 
(1) reducing the input burden from complete, noise-free shape representations to point clouds, which significantly expands potential use cases, (2) providing fast inference and scalable training given any cohort size, (3) allowing for partial input via simultaneous SSM prediction and completion, (4) enabling sequential or online learning, as well as incremental model updating as clinical studies progress, and (5) eliminating biases introduced by metrics and parametric representations used in classical methods.
By enabling SSM from point clouds, we can increase SSM accessibility and potentially accelerate its adoption as a widespread clinical tool.

\subsubsection{Acknowledgements}
This work was supported by the National Institutes of Health under grant numbers NIBIB-U24EB029011, NIAMS-R01AR076120, \\ NHLBI-R01HL135568, and  NIBIB-R01EB016701.
The content is solely the responsibility of the authors and does not necessarily represent the official views of the National Institutes of Health.
The authors would like to thank the University of Utah Division of Cardiovascular Medicine for providing left atrium MRI scans and segmentations from the Atrial Fibrillation projects as well as the
Orthopaedic Research Laboratory (Andrew Anderson, PhD) at the University of Utah for providing femur CT scans and corresponding segmentations.
%


\bibliographystyle{splncs04}
\bibliography{ref}

\appendix
\section{Dataset Visualization}
\begin{figure}
\centering
    \includegraphics[width=\textwidth]{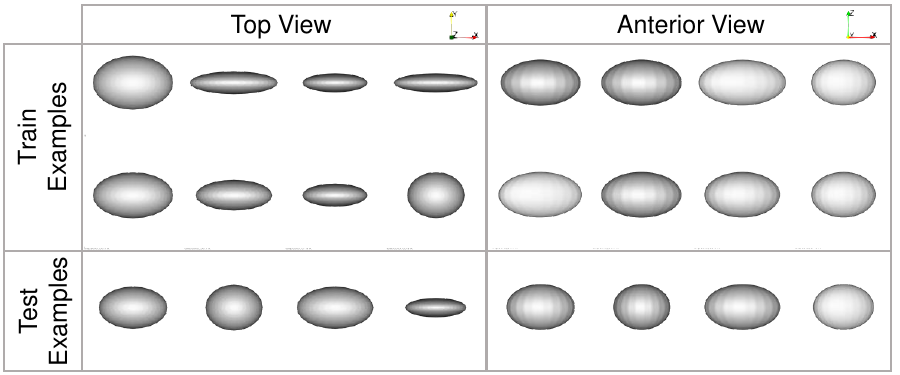}
    \caption{\textbf{Ellipsoid Dataset} Eight of the 50 training examples and 4 of the 30 test examples are shown from both the top and anterior view. Ellipsoids have random $x$ and $y$-radius and fixed $z$ radius.}
    \label{fig:app_ellip}
\end{figure}

\begin{figure}
    \centering
    \includegraphics[width=\textwidth]{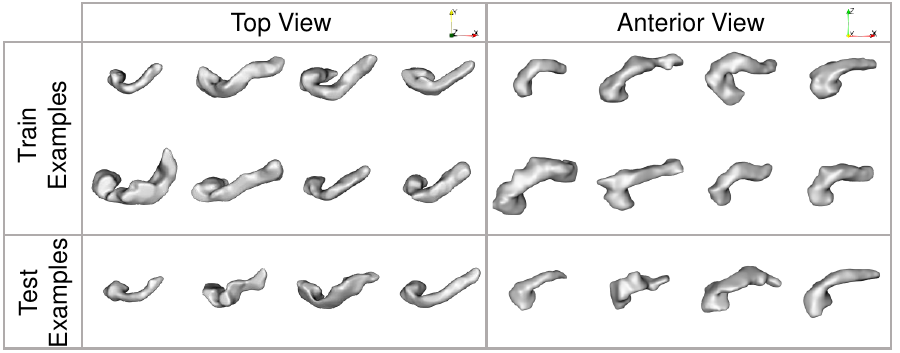}
    \caption{\textbf{Pancreas Dataset \cite{simpson2019medseg}} Eight of the 245 training examples and 4 of the 28 test examples are shown from both top and anterior view.}
    \label{fig:app_panc}
\end{figure}

Figures \ref{fig:app_ellip}-\ref{fig:app_panc} display example shapes from each of the datasets used in experiments from two viewpoints. 

\begin{figure}
    \centering
    \includegraphics[width=\textwidth]{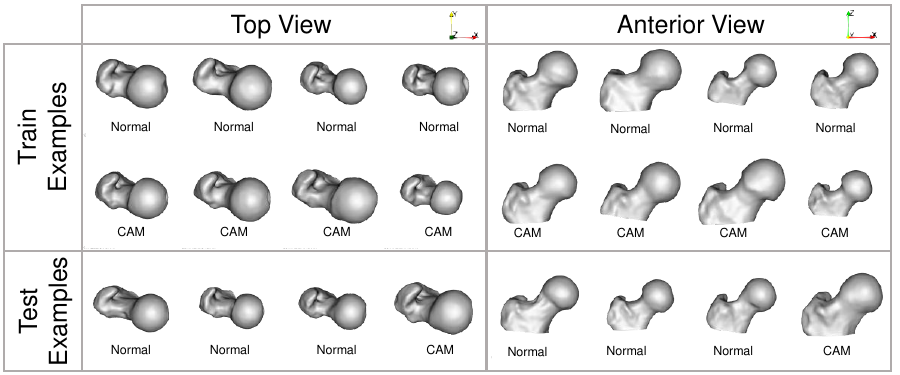}
    \caption{\textbf{Femur Dataset} Eight of the 51 training examples and 4 of the 5 test examples are shown from both top and anterior view. The group label is displayed, denoting which femurs have the CAM pathology. Overall 47 of the femurs were normal and 9 were CAM, 8 were in the train set, and 1 was in the test.}
\end{figure}

\begin{figure}
    \centering
    \includegraphics[width=\textwidth]{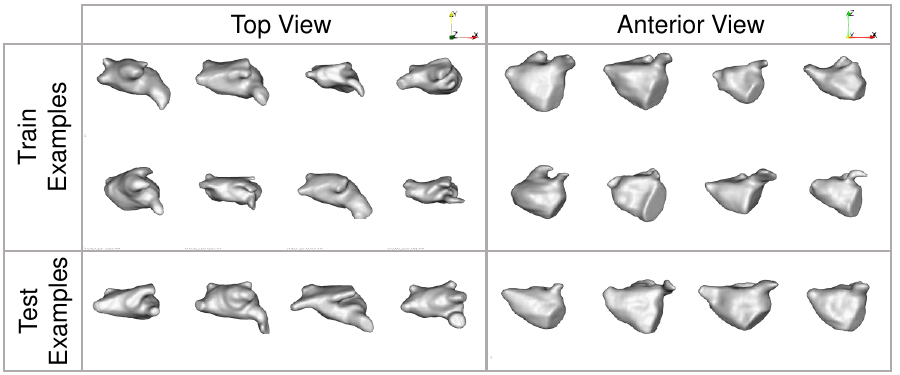}
    \caption{\textbf{Left Atrium Dataset} Eight of the 987 training examples and 4 of the 109 test examples are shown from both top and anterior view.}
\end{figure}

\begin{figure}
    \centering
    \includegraphics[width=\textwidth]{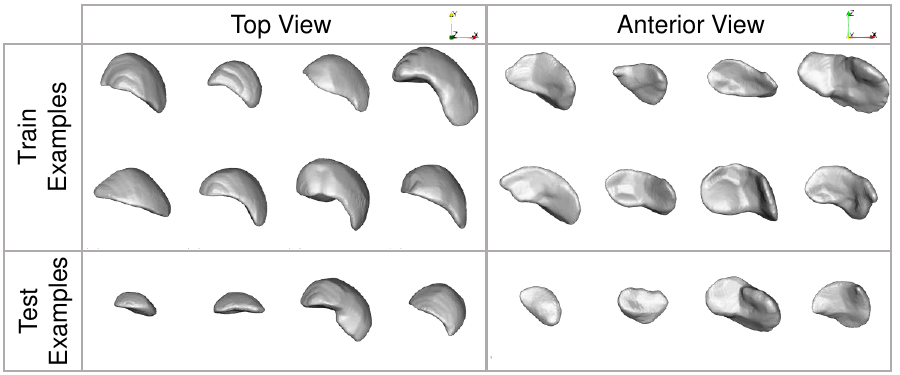}
    \caption{\textbf{Spleen Dataset \cite{simpson2019medseg}} Eight of the 36 training examples and all 4 of the test examples are shown from both top and anterior view.}
\end{figure}

\end{document}